\documentclass[sigconf]{acmart}

\AtBeginDocument{%
  }

\copyrightyear{2022}
\acmYear{2022}
\setcopyright{acmcopyright}\acmConference[KDD '22]{Proceedings of the 28th ACM SIGKDD Conference on Knowledge Discovery and Data Mining}{August 14--18, 2022}{Washington, DC, USA}
\acmBooktitle{Proceedings of the 28th ACM SIGKDD Conference on Knowledge Discovery and Data Mining (KDD '22), August 14--18, 2022, Washington, DC, USA}
\acmPrice{15.00}
\acmDOI{10.1145/3534678.3539300}
\acmISBN{978-1-4503-9385-0/22/08}

\usepackage[utf8]{inputenc} 
\usepackage[T1]{fontenc}    
\usepackage{hyperref}       
\usepackage{url}            
\usepackage{booktabs}       
\usepackage{amsfonts}       
\usepackage{nicefrac}       
\usepackage{microtype}      
\usepackage{xcolor}         

\usepackage{booktabs}       
\usepackage{amsfonts}       
\usepackage{microtype}      
\usepackage{url}
\usepackage{verbatim} 
\usepackage{graphicx}
\usepackage{multirow}
\usepackage{algorithm}
\usepackage[noend]{algorithmic}
\usepackage{xspace}
\usepackage{epsfig}
\usepackage{amsmath}
\usepackage{amsthm}
\usepackage{xr}
\usepackage{bbm}
\usepackage{bm}
\usepackage{subcaption}
\usepackage{enumitem}
\usepackage{hyperref}       
\usepackage{multicol}
\usepackage{tabularx}
\usepackage{perpage} 
\MakePerPage{footnote} 
\usepackage{wrapfig}

\newcommand{\xhdr}[1]{{\noindent\bfseries #1}.}

\newcommand{\mb}{\mathbf}
\newcommand{\cut}[1]{}

\newcommand{\eg}{\emph{e.g.}}

\newcommand{\name}{ROLAND\xspace}

\usepackage{amsmath,amsfonts,bm}

\def\eqref#1{equation~\ref{#1}}

\def\1{\bm{1}}

\DeclareMathAlphabet{\mathsfit}{\encodingdefault}{\sfdefault}{m}{sl}
\SetMathAlphabet{\mathsfit}{bold}{\encodingdefault}{\sfdefault}{bx}{n}

\usepackage{cleveref}
\usepackage{adjustbox}
\usepackage{soul}

\begin{document}

\title{
ROLAND: Graph Learning Framework for Dynamic Graphs
}


\author{Jiaxuan You}
\email{jiaxuan@cs.stanford.edu}
\affiliation{
\institution{Stanford University}
\state{California}
\country{USA}
}

\author{Tianyu Du}
\email{tianyudu@stanford.edu}
\affiliation{
\institution{Stanford University}
\state{California}
\country{USA}
}

\author{Jure Leskovec}
\email{jure@cs.stanford.edu}
\affiliation{
\institution{Stanford University}
\state{California}
\country{USA}
}

\begin{abstract}

Graph Neural Networks (GNNs) have been successfully applied to many real-world static graphs.
However, the success of static graphs has not fully translated to dynamic graphs due to the limitations in model design, evaluation settings, and training strategies.
Concretely, existing dynamic GNNs do not incorporate state-of-the-art designs from static GNNs, which limits their performance.
Current evaluation settings for dynamic GNNs do not fully reflect the evolving nature of dynamic graphs.
Finally, commonly used training methods for dynamic GNNs are not scalable.
Here we propose ROLAND, an effective graph representation learning framework for real-world dynamic graphs.
At its core, the ROLAND framework can help researchers easily repurpose any static GNN to dynamic graphs. Our insight is to view the node embeddings at different GNN layers as hierarchical node states and then recurrently update them over time.
We then introduce a live-update evaluation setting for dynamic graphs that mimics real-world use cases, where GNNs are making predictions and being updated on a rolling basis.
Finally, we propose a scalable and efficient training approach for dynamic GNNs via incremental training and meta-learning.
We conduct experiments over eight different dynamic graph datasets on future link prediction tasks.
Models built using the ROLAND framework achieve on average 62.7\% relative mean reciprocal rank (MRR) improvement over state-of-the-art baselines under the standard evaluation settings on three datasets. 
We find state-of-the-art baselines experience out-of-memory errors for larger datasets, while ROLAND can easily scale to dynamic graphs with 56 million edges.
After re-implementing these baselines using the ROLAND training strategy, ROLAND models still achieve on average 15.5\% relative MRR improvement over the baselines.

\end{abstract}



\maketitle

\section{Introduction and Related Work}

The problem of learning from dynamic networks arises in many application domains, such as fraud detection \cite{khazane2019deeptrax,Li2019ClassifyingAU}, anti-money laundering \cite{weber2019anti}, and recommender systems \cite{you2019hierarchical}.
Graph Neural Networks (GNNs) are a general class of models that can perform various learning tasks on graphs.
GNNs have gained tremendous success in learning from static graphs \cite{battaglia2018relational,hamilton2017inductive,zhang2018link,you2019position,you2021identity}. Although various GNNs have been proposed for dynamic graphs \cite{chen2018gc,li2019predicting,li2017diffusion,pareja2020evolvegcn,peng2020spatial,seo2018structured,taheri2019learning,wang2020traffic,yu2017spatio,zhao2019t} 
these approaches have limitations of \emph{model design}, \emph{evaluation settings} and \emph{training strategies}.
Overcoming these limitations is crucial for real-world dynamic graph applications.

\begin{figure*}[t]
\centering
\vspace{-2mm}
\includegraphics[width=\linewidth]{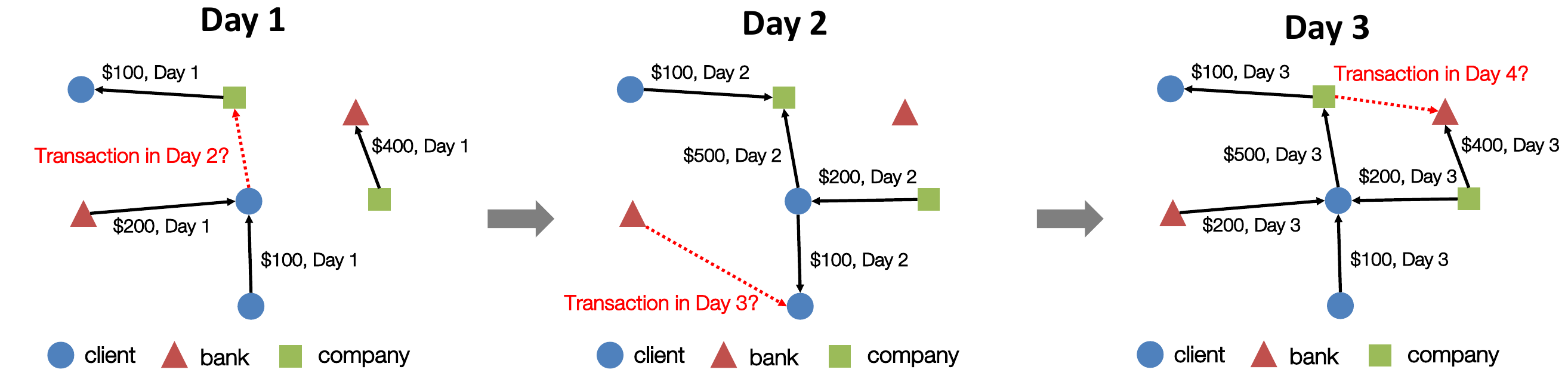} 
\vspace{-6mm}
\caption{\textbf{Example ROLAND use case} for future link prediction on a dynamic transaction graph. We use information up to time $t$ to predict potential edges at time $t+1$.}

\label{fig:overview}
\end{figure*}

\xhdr{Limitations of model design}
Existing approaches fail to transfer successful static GNN designs/architectures to dynamic graph applications.
Many existing works treat a GNN as a feature encoder and then build a sequence model on top of the GNN \cite{peng2020spatial, wang2020traffic, yu2017spatio}.
Other works take Recurrent Neural Networks (RNNs), then replace the linear layers in the RNN cells with graph convolution layers \cite{li2018dcrnn_traffic, seo2018structured, zhao2019t}.
Although these approaches are intuitive, they do not incorporate state-of-the-art designs from static GNNs.
For instance, the incorporation of skip-connections \cite{dwivedi2020benchmarking,he2016deep,li2021deepgcns}, batch normalization \cite{dwivedi2020benchmarking,ioffe2015batch,you2020design}, edge embedding \cite{gilmer2017neural,you2020handling} has been beneficial for GNN message passing, but has not been explored for dynamic GNNs.
To avoid re-exploring these design choices for dynamic GNNs, instead of building dynamic GNNs from scratch, a better design principle would be to start from a mature static GNN design and adapt it for dynamic graphs.

\xhdr{Limitations of evaluation settings}
When evaluating dynamic GNNs, existing literature usually ignores the evolving nature of data and models.
Concretely, dynamic graph datasets are often deterministically split by time, \eg, if ten months of data are available, the first eight months of data will be used as training, one month as validation, and the last month as the test set \cite{pareja2020evolvegcn,zhao2019t}.
Such protocol evaluates the model only on edges from the last month of the available dataset; therefore, it tends to overestimate the model performance given the presence of long-term pattern changes.
Additionally, in most literature, a non-updated model is used for prediction within the time span at evaluation time, which means the model gets stale over time \cite{li2017diffusion, wang2020traffic, zhao2019t}. 
However, in real applications, users can update their models based on new data and then make future predictions.

\xhdr{Limitations of training strategies}
The common training methods for dynamic GNNs can be improved in terms of scalability and efficiency \cite{seo2018structured,sharma2020forecasting,zhao2019t}.
First, most existing training methods for dynamic GNNs usually require keeping the entire or a large portion of the graph in GPU memory \cite{seo2018structured,sharma2020forecasting,zhao2019t}, since all or a large portion of historical edges are used for message passing.
Therefore, dynamic GNNs are often evaluated on small networks with only a few hundred nodes \cite{diao2019dynamic, yu2017spatio} or small transaction graphs with fewer than 2 million edges \cite{pareja2020evolvegcn, sharma2020forecasting}.
Moreover, there is little research on making dynamic GNNs generalize and quickly adapt to new data.

\xhdr{Present work}
Here we propose \name, an effective graph representation learning framework for real-world dynamic graphs. We focus on the snapshot-based representation of dynamic graphs where nodes and edges are arriving in batches (\eg, daily, weekly).
\name framework can help researchers re-purpose any static GNN to a dynamic graph learning task; consequently, we can adapt state-of-the-art designs from static GNNs and significantly lower the barrier to learning from dynamic graphs.

\xhdr{ROLAND model}
Our insight is that any GNN for static graphs can be extended for dynamic graph use cases. 
We offer a new viewpoint for static GNNs, where the node embeddings at different GNN layers are viewed as \emph{hierarchical node states}.
To generalize a static GNN to a dynamic setting, we only need to define how to update these hierarchical nodes states based on newly observed nodes and edges.
We explore a variety of update-modules, including moving average, Multi-layer Perceptron (MLP), and Gated Recurrent Unit (GRU) \cite{chung2014empirical}.
This way of building dynamic GNNs is simple and effective, and more importantly, this approach can keep the design of a given static GNN and use it on a dynamic graph.

\xhdr{ROLAND evaluation}
We then introduce a live-update evaluation setting for dynamic graphs, where GNNs are used to make predictions on a rolling basis (\eg, daily, weekly). When making each prediction, we make sure the model has only been trained using historical data so that there is no information leakage from the future to the past. We additionally allow the model to be fine-tuned using the new graph snapshot, mimicking real-world use cases, in which the model is constantly updated to fit evolving data.

\xhdr{ROLAND training}
Finally, we propose a scalable and efficient training approach for dynamic GNNs.
We propose an incremental training strategy that does a truncated version of back-propagation-through-time (BPTT) \cite{williams1990efficient}, which significantly saves GPU memory cost.
Concretely, we only keep the incoming new graph snapshot and historical node states in GPU memory.
Using this technique, we are able to train GNNs on a dynamic transaction network with 56 million edges, which is about 13 times larger than existing benchmarks in terms of edges per snapshot.
Furthermore, we formulate prediction tasks over dynamic graphs as a \emph{meta-learning problem}, where we treat making predictions in different periods (\eg, every day) as different tasks arriving sequentially.
We find a meta-model, which serves as a good initialization that is used to derive a specialized model for future unseen prediction tasks quickly.

\xhdr{Key results}
We conduct experiments over eight different dynamic graph datasets, with up to 56 million edges and 733 graph snapshots.
We first compare \name to six baseline models under the evaluation setting in the existing literature on three datasets, \name achieves 62.7\% performance gain over the best baseline on average, demonstrating the effectiveness of \name design framework.
We then evaluate our models and baselines on the proposed live-update setting, which provides a more realistic evaluation setting for dynamic graphs.
We find that existing dynamic GNN training methods based on BPTT fail to scale to large datasets.
To get meaningful comparisons, we re-implement baselines using \name training, which greatly reduces GPU memory cost.
\name still achieves on average 15.5\% performance gain over the \name version of baselines.

\xhdr{Contributions}
We summarize ROLAND's innovation as follows.
\begin{enumerate}
    \item \emph{Model design}: ROLAND describes how to repurpose a static GNN for dynamic settings effectively.
    Furthermore, ROLAND shows that successful static GNN designs lead to significant performance gain on dynamic prediction tasks.
    \item \emph{Training}: ROLAND can scale to dynamic graphs with 56 million edges, which is at least 13 times larger than existing benchmarks.
    In addition, we innovatively formulate predicting over dynamic graphs as a meta-learning problem to achieve fast model adaptation. 
    \item \emph{Evaluation}: Compared to the common deterministic dataset split, ROLAND’s live-update evaluation can reflect the evolving nature of data and model.
\end{enumerate}

\section{Preliminaries}

\begin{figure*}[t]
\centering
\vspace{-1mm}
\includegraphics[width=\linewidth]{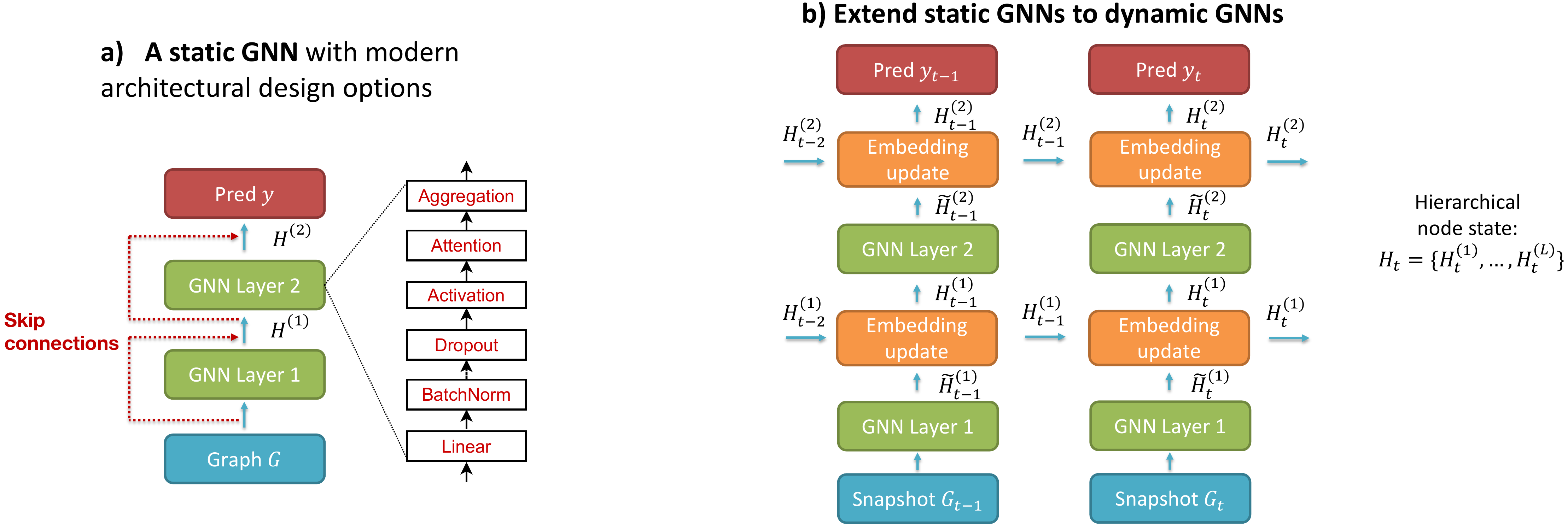} 
\vspace{-4mm}
\caption{\textbf{ROLAND model design principle}. \textbf{(a)} Example static GNN with modern architecture designs, including intra-layer designs such as BatchNorm and attention, and inter-layer designs such as skip-connections. \textbf{(b)} ROLAND can easily extend any static GNN to dynamic graphs, by inserting embedding update modules that update hierarchical node states $H_t$ over time. }
\label{fig:overview}
\end{figure*}

\label{subsec:preliminaries}

\xhdr{Graphs and dynamic graphs}
A graph can be represented as $G = (V,E)$, where $V = \{v_1, ..., v_n\}$ is the node set and $E \subseteq V \times V$ is the edge multi-set. Nodes can be paired with features $X = \{\mb{x}_v \mid v \in V\}$, and edges can also have features $F = \{\mb{f}_{uv} \mid (u,v) \in E\}$. 
For dynamic graphs, each node $v$ further has a timestamp $\tau_v$ and edge $e$ has a timestamp $\tau_e$.
We focus on the snapshot-based representation for dynamic graphs: a dynamic graph $\mathcal{G}=\{G_t\}_{t=1}^T$ can be represented as a sequence of graph snapshots, where each snapshot is a static graph $G_t = (V_t, E_t)$ with $V_t = \{v\in V \mid \tau_v = t\}$ and $E_t = \{e \in E \mid \tau_e = t\}$.
By modeling over graph snapshots, ROLAND can naturally handle node addition/deletion since different graph snapshots could have different sets of nodes.

\xhdr{Graph Neural Networks}
The goal of a GNN is to learn node embeddings based on an iterative aggregation of messages from the local network neighborhood. We use embedding matrix $H^{(L)} = \{\mb{h}_v^{(L)}\}_{v \in V}$ to denote the embedding for all the nodes after applying an $L$-layer GNN. The $l$-th layer of a GNN, $H^{(l)} = \textsc{Gnn}^{(l)}(H^{(l-1)})$, can be written as:
\begin{equation}
    \begin{aligned}
    &\mb{m}_{u \to v}^{(l)} = \textsc{Msg}^{(l)}(\mb{h}_u^{(l-1)}, \mb{h}_{v}^{(l-1)}) \\
    &\mb{h}_v^{(l)} = \textsc{Agg}^{(l)}\big(\{\mb{m}_{u \to v}^{(l)} \mid u \in \mathcal{N}(v)\}, \mb{h}_v^{(l-1)}\big)
    \end{aligned}
    \label{eq:gnn}
\end{equation}
where $\mb{h}_v^{(l)}$ is the node embedding for $v \in V$ after passing through $l$ layers , $\mb{h}_v^{(0)}=\mb{x}_v$, $\mb{m}_v^{(l)}$ is the message embedding, and $\mathcal{N}(v)$ is the direct neighbors of $v$. Different GNNs can have have various definitions of message-passing functions $\textsc{Msg}^{(l)}(\cdot)$ and aggregation functions $\textsc{Agg}^{(l)}(\cdot)$.
We refer to GNNs defined in Equation (\ref{eq:gnn}) as static GNNs, since they are designed to learn from static graphs and do not capture the dynamic information of the graph.

\section{Proposed ROLAND Framework}

\subsection{From Static GNNs to Dynamic GNNs}
Here we propose the \name framework that generalizes static GNNs to a dynamic setting, illustrated in Figure \ref{fig:overview}. We summarize the forward computation of the ROLAND model in Algorithm \ref{alg:gnn-forward}.

\xhdr{Revisit static GNNs}
A static GNN often consists of a stack of $L$ GNN layers, where each layer follows the general formulation in Equation (\ref{eq:gnn}).
We denote the node embedding matrix after the $l$-th GNN layer as 
$H^{(l)} = \{\mb{h}_v^{(l)}\}_{v \in V}$.
The node embeddings computed in all layers characterize a given node in a hierarchical way; concretely, $H^{(l)}$ summarizes the information from the neighboring nodes that are $l$ hops away from a given node. We use $H=\{H^{(1)},...,H^{(L)}\}$ to represent the embeddings in all GNN layers.

\xhdr{From static embeddings to dynamic states}
We propose to extend the semantics of $H$ from static embeddings to dynamic node states. Building on previous discussions, we view $H_t$ as the \emph{hierarchical node state} at time $t$, where each $H^{(l)}$ captures multi-hop node neighbor information.
Suppose the input graph $G$ has dynamically changed, then the computed embedding $H$ will also change. 
Therefore, we argue that the key to generalizing any static GNN to a dynamic graph relies on how to update the hierarchical node states $H$ over time.

\xhdr{Differences with prior works}
In the common approach where a sequence model is built on top of a GNN, only the top-level node state $H^{(L)}$ is being kept and updated \cite{peng2020spatial, wang2020traffic, yu2017spatio}. All the lower-level node states $H^{(1)}_{t}, ..., H^{(l)}_{t}$ are always recomputed from scratch based on the new graph snapshot $G_t$. Here we propose to keep and update the entire hierarchical node state $H^{(1)}_{t}, ..., H^{(l)}_{t}$ at all levels.

\begin{algorithm}[t]
\caption{ROLAND GNN forward computation}
\label{alg:gnn-forward}
\begin{flushleft}
\textbf{Input:} Dynamic graph snapshot $G_t$, hierarchical node state $H_{t-1}$
\textbf{Output:} Prediction $y_t$, updated node state $H_t$
\end{flushleft}

\begin{algorithmic}[1]
\STATE $H_t^{(0)} \leftarrow X_t$ \hfill \COMMENT{Initialize embedding from $G_t$}
\FOR{$l = 1,\dots,L$}
\STATE $\tilde{H}_t^{(l)} = \textsc{Gnn}^{(l)}(H_t^{(l-1)})$ \hfill\COMMENT{Implemented as Equation (\ref{eq:gnn_roland})}
\STATE $H^{(l)}_{t} = \textsc{Update}^{(l)}(H^{(l)}_{t-1}, \tilde{H}_t^{(l)})$ \hfill\COMMENT{Equation (\ref{eq:update})}
\ENDFOR
\STATE $y_t=\textsc{Mlp}(\textsc{Concat}(\mb{h}_{u, t}^{(L)}, \mb{h}_{v, t}^{(L)})), \forall (u, v) \in E$ \hfill\COMMENT{Equation (\ref{eq:pred_head})}
\end{algorithmic}
\end{algorithm}

\xhdr{Hierarchically update modules in dynamic GNNs}
In ROLAND, we propose an \emph{update-module} that updates node embeddings hierarchically and dynamically. The update-module can be inserted to any static GNN. The new level $l$ node state $H^{(l)}_{t}$ depends on both lower layer node state $\tilde{H}_t^{(l)}$ and historical node state $H_{t-1}^{(l)}$.
\begin{align}
\label{eq:update}
    \hspace{-2mm} H^{(l)}_{t} = \textsc{Update}^{(l)}(H^{(l)}_{t-1},\tilde{H}_{t}^{(l)}),\hspace{2mm} \tilde{H}_{t}^{(l)} = \textsc{Gnn}^{(l)}(H_{t}^{(l-1)})
\end{align}
We examine three simple yet effective embedding update methods.
(1) \emph{Moving Average}: embedding of node $v$ at time $t$ is updated by
$H^{(l)}_{t,v} = \kappa_{t, v} H^{(l)}_{t-1, v} + (1-\kappa_{t,v}) \tilde{H}^{(l)}_{t,v}$. Moving average can naturally capture dynamics of embedding and is free from trainable parameters.
We define the moving average weight $\kappa_{t,v}$ as
\begin{align}
    \kappa_{t,v} = 
    \frac{\sum_{\tau = 1}^{t-1} |E_\tau|}{\sum_{\tau = 1}^{t-1} |E_\tau| + |E_t|} \in [0, 1]
    \label{eq:keep-ratio}
\end{align}
(2) \emph{MLP}: node embeddings are updated by a 2-layer MLP, $H^{(l)}_{t} = \textsc{Mlp}(\textsc{Concat}(H^{(l)}_{t-1} ,\tilde{H}_t^{(l-1)}))$. 
(3) \emph{GRU}: node embeddings are updated by a GRU cell \cite{chung2014empirical}, $H^{(l)}_{t} = \textsc{Gru}(H^{(l)}_{t-1}, \tilde{H}_t^{(l)})$, where $H^{(l)}_{t-1}$ is the hidden state and $\tilde{H}_t^{(l)}$ is the input for the GRU cell.

\xhdr{GNN architecture design in ROLAND}
In ROLAND, we extend the GNN layer definition in Equation (\ref{eq:gnn}) to incorporate successful GNN designs in static GNNs. Specifically, we have
\begin{equation}
    \begin{aligned}
    & \mb{m}_{u\to v}^{(l)} = \mb{W}^{(l)}\textsc{Concat}\big(\mb{h}_u^{(l-1)}, \mb{h}_v^{(l-1)}, \mb{f}_{uv}\big), \\
    & \mb{h}_v^{(l)} = \textsc{Agg}^{(l)}\big(\{\mb{m}_{u\to v}^{(l)} \mid u \in \mathcal{N}(v)\}\big) + \mb{h}_v^{(l-1)}
    \end{aligned}
    \label{eq:gnn_roland}
\end{equation}
Here, we consider edge features $\mb{f}_{uv}$ in message computation $\textsc{Msg}^{(l)}(\cdot)$, since dynamic graphs usually have timestamps as edge features; additionally, we include $\mb{h}_v^{(l-1)}$ in the message passing computation, so that bidirectional information flow can be modeled; we explore different aggregation functions $\textsc{Agg}$ including summation $\textsc{Sum}$, maximum $\textsc{Max}$, and average $\textsc{Mean}$; finally, we add skip-connections \cite{he2016deep,li2021deepgcns} when stacking GNN layers by adding $\mb{h}_v^{(l-1)}$ after message aggregation. We show in Section \ref{subsec:ablation} that these designs lead to performance boost.
Based on the computed node embeddings, \name predicts the probability of a future edge from node $u$ to $v$ via an MLP prediction head
\begin{equation}
\label{eq:pred_head}
y_t=\textsc{Mlp}(\textsc{Concat}(\mb{h}_{u, t}^{(L)}, \mb{h}_{v, t}^{(L)})), \forall (u, v) \in V \times V
\end{equation}

\begin{figure}[t]
\centering
\includegraphics[width=0.98\linewidth]{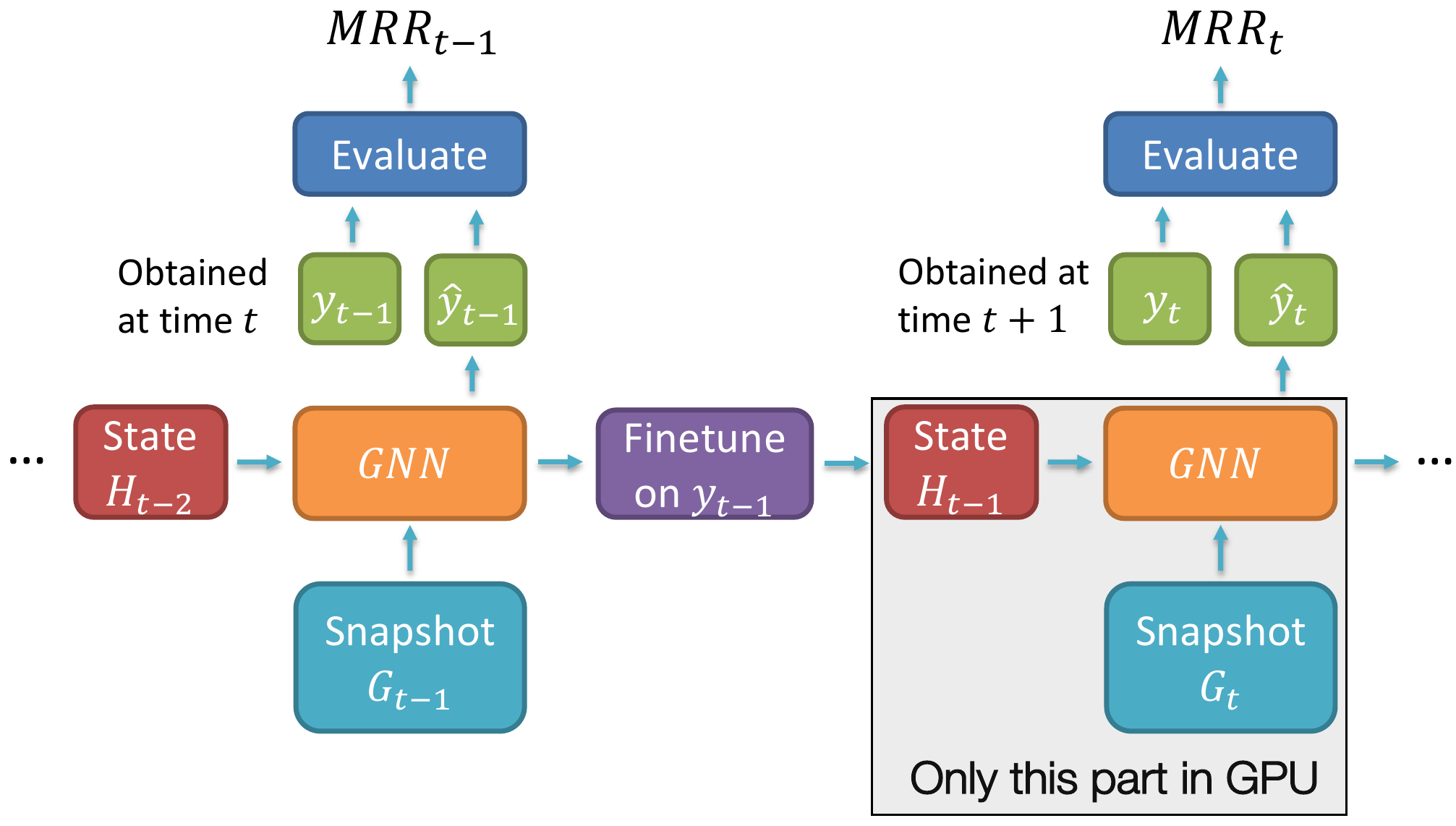} 
\vspace{-1mm}
\caption{\textbf{ROLAND live-update evaluation}. ROLAND fine-tunes $\textsc{Gnn}$ with $y_{t-1}$ and updates node embeddings $H_{t-1}$ for the next prediction task. After obtaining labels $y_{t}$, ROLAND evaluates $\textsc{Gnn}$'s predictive performance based on historical state $H_{t-1}$ and current snapshot $G_{t}$. 
}
\label{fig:eval}
\end{figure}

\subsection{Live-update Evaluation}
\vspace{-1mm}

\xhdr{Standard fixed split evaluation setting}
The evaluation procedure used in prior works constructs training and testing sets by splitting available graph snapshots sequentially. For example, the first 90\% of snapshots for training and cross-validation, then models are evaluated using all edges from the last 10\% of snapshots \cite{pareja2020evolvegcn}.
However, the data distribution of dynamic graphs is constantly evolving in the real world. For example, the number of purchase transactions made in the holiday season is higher than usual.
Therefore, evaluating models solely based on edges from the last 10\% of snapshots can provide misleading results.

\xhdr{Live-update evaluation setting}
To utilize all snapshots to evaluate the model, we propose a live-update evaluation procedure (Figure \ref{fig:eval} and Algorithm \ref{alg:live-update-evaluation}), which consists of the following key steps:
(1) Fine-tune GNN model using newly observed data
(2) evaluate predictions using new data,
(3) making predictions using historical information,
Specifically, ROLAND first collects link prediction labels $y_{t-1}$ at time $t$ (predict future links), which are split into $y_{t-1}^{(train)}$ and $y_{t-1}^{(val)}$. Then, ROLAND fine-tunes $\textsc{Gnn}$ on training labels $y_{t-1}^{(train)}$, while performance $\textsc{MRR}_{t-1}^{(val)}$ on validation labels $y_{t-1}^{(val)}$ are used as the early stopping criterion. We use Mean reciprocal rank (MRR) instead of ROC AUC, since negative labels significantly outnumber positive labels in a link prediction task, and MMR has been adopted in prior work as well \cite{pareja2020evolvegcn}.
Finally, after obtaining edge labels $y_{t}$ at time $t+1$, we report ROLAND's performance $\textsc{MRR}_{t}$ based on historical state $H_{t-1}$ and current snapshot $G_{t}$. This live-update evaluation is free from information leakage, since no future information is used during both training and evaluation.
We report the average $\textsc{MRR}_{t}$ over all prediction steps as the final model performance.

\begin{algorithm}[t]
\caption{ROLAND live-update evaluation}
\label{alg:live-update-evaluation}
\begin{flushleft}
\textbf{Input:} Dynamic graph $\mathcal{G}=\{G_1,\dots,G_T\}$, link prediction labels $y_1,\dots,y_T$, number of snapshots $T$, $\textsc{Gnn}(\cdot)$ defined in Algorithm \ref{alg:gnn-forward}\\
\textbf{Output:} Performance $\textsc{MRR}$, model $\textsc{Gnn}$
\end{flushleft}

\begin{algorithmic}[1]
\STATE Initialize hierarchical node state $H_0$
\FOR{$t = 2,\dots,T$}
\STATE Collect link prediction labels $y_{t-1} = y_{t-1}^{(train)} \cup y_{t-1}^{(val)}$, $y_{t}$
\WHILE{$\textsc{MRR}_{t-1}^{(val)}$ is increasing}
\STATE $H_{t-1}, \hat{y}_{t-1} \leftarrow \textsc{Gnn}(G_{t-1}, H_{t-2})$, $\hat{y}_{t-1} = \hat{y}_{t-1}^{(train)} \cup \hat{y}_{t-1}^{(val)}$
\STATE Update $\textsc{Gnn}$ via backprop based on $\hat{y}_{t-1}^{(train)}$, $y_{t-1}^{(train)}$
\STATE $\textsc{MRR}_{t-1}^{(val)} \leftarrow \textsc{Evaluate}(\hat{y}_{t-1}^{(val)}, y_{t-1}^{(val)})$
\ENDWHILE
\STATE $H_{t}, \hat{y}_{t} \leftarrow \textsc{Gnn}(G_{t}, H_{t-1})$
\STATE $\textsc{MRR}_{t} \leftarrow \textsc{Evaluate}(\hat{y}_{t}, y_{t})$
\ENDFOR
\STATE $\textsc{MRR}=\sum_{t=2}^{T} \textsc{MRR}_t/(T-1)$
\end{algorithmic}
\end{algorithm}

\begin{table*}[t]
\centering
\caption{\textbf{Summary of dataset statistics}.}
\vspace{-1mm}
\begin{footnotesize}
\resizebox{0.8\textwidth}{!}{
\begin{tabular}{cccccc}
\toprule
 & \# Edges & \# Nodes & Range & Snapshot Frequency & \# Snapshots \\
\midrule
BSI-ZK & 56,194,191 & 1,744,561 & Jan 01, 2008 - Dec 30, 2008 & daily & 257 \\
AS-733 & 11,965,533 & 7,716 & Nov 8, 1997 - Jan 2, 2000 & daily & 733 \\
Reddit-Title & 571,927 & 54,075 & Dec 31, 2013 - Apr 30, 2017 & weekly & 178 \\
Reddit-Body & 286,561 & 35,776 & Dec 31, 2013 - Apr 30, 2017 & weekly & 178 \\
BSI-SVT & 190,133 & 89,564 & Jan 27, 2008 - Dec 30, 2008 & weekly & 49 \\
UCI-Message & 59,835 & 1,899 & Apr 15, 2004 - Oct 26, 2004 & weekly & 29 \\
Bitcoin-OTC & 35,592 & 5,881 & Nov 8, 2010 - Jan 24, 2016 & weekly & 279 \\
Bitcoin-Alpha & 24,186 & 3,783 & Nov 7, 2010 - Jan 21, 2016 & weekly & 274 \\
\bottomrule
\end{tabular}
}
\label{tab:datasets}
\end{footnotesize}
\end{table*}

\subsection{Training Strategies}

\xhdr{Scalability: Incremental training}
We propose a scalable and efficient training approach for dynamic GNNs that borrows the idea of the truncated version of back-propagation-through-time, which has been widely used for training RNNs \cite{williams1990efficient}.
Concretely, we only keep the model $\textsc{Gnn}$, the incoming new graph snapshot $G_t$, and historical node states $H_{t-1}$ into GPU memory.
Since the historical node states $H_{t-1}$ has already encoded information up to time $t-1$, instead of training $\textsc{Gnn}$ to predict $y_t$ from the whole sequence $\{G_1, \dots, G_{t-1}\}$, the model only needs to learn from $\{H_{t-1}$, and the newly observed graph snapshot $G_{t}\}$. With this incremental training strategy, ROLAND's memory complexity is agnostic to the number of graph snapshots, and we can train GNNs on dynamic graphs with 56 million edges and 733 graph snapshots.

\begin{algorithm}[t]
\caption{ROLAND training algorithm}
\label{alg:train}
\begin{flushleft}
\textbf{Input:} Graph snapshot $G_t$, link prediction label $y_t$, hierarchical node state $H_{t-1}$, smoothing factor $\alpha$, meta-model $\textsc{Gnn}^{(meta)}$\\
\textbf{Output:} Model $\textsc{Gnn}$, updated meta-model $\textsc{Gnn}^{(meta)}$
\end{flushleft}

\begin{algorithmic}[1]
\STATE $\textsc{Gnn} \leftarrow \textsc{Gnn}^{(meta)}$
\STATE Move $\textsc{Gnn}, G_{t}, H_{t-1}$ to GPU 
\WHILE{$\textsc{MRR}_t^{(val)}$ is increasing}
\STATE $H_t, \hat{y}_t \leftarrow \textsc{Gnn}(G_t, H_{t-1})$, $\hat{y}_t = \hat{y}_t^{(train)} \cup \hat{y}_t^{(val)}$
\STATE Update $\textsc{Gnn}$ via backprop based on $\hat{y}_t^{(train)}$, $y_t^{(train)}$
\STATE $\textsc{MRR}_t^{(val)} \leftarrow \textsc{Evaluate}(\hat{y}_t^{(val)}, y_t^{(val)})$
\ENDWHILE
\STATE Remove $\textsc{Gnn}, G_{t}, H_{t-1}$ from GPU
\STATE $\textsc{Gnn}^{(meta)} \leftarrow  (1-\alpha)\textsc{Gnn}^{(meta)} + \alpha\textsc{Gnn}$
\end{algorithmic}
\end{algorithm}

\xhdr{Fast adaptation: Meta-training}
We formulate prediction tasks over dynamic graphs as a meta-learning problem, where we treat making predictions in different periods (\eg, every day) as different tasks.
The standard approach of updating $\textsc{Gnn}$ is simply fine-tuning the model using new information.
However, always keeping the prior model for the next future prediction task is not necessarily optimal.
For example, for dynamic graphs with weekly patterns, the optimal model on Friday may be drastically different from the optimal model on Saturday; therefore, simply fine-tuning the previous model could lead to inferior models.
Instead, our proposal is to find a \emph{meta-model} $\text{GNN}^{(meta)}$, which serves as a good initialization that is used to derive a specialized model for future unseen prediction tasks quickly. In principle, any meta-learning algorithm can be used to update $\text{GNN}^{(meta)}$; we follow the Reptile algorithm \cite{nichol2018first} which is simple and effective. As is shown in Algorithm \ref{alg:train}, at each time $t$, we first initialize the $\text{GNN}$ model using $\text{GNN}^{(meta)}$ and then use back-propagation with early stopping to fine-tune the model for the next prediction task.
Then, we updates the meta-model $\text{GNN}^{(meta)}$ by computing a moving average of the trained models $(1-\alpha)\textsc{Gnn}^{(meta)} + \alpha\textsc{Gnn}$, where $\alpha \in [0, 1]$ is the smoothing factor.

\begin{table*}[ht]
\setlength\tabcolsep{3pt}
\centering
\caption{\textbf{Results in the standard fixed split setting.} Performance of the baselines are reported in the EvolveGCN paper \cite{pareja2020evolvegcn}. Our experiments ensure the same snapshot frequency, set of test snapshots and method to compute MRRs, etc., for fair comparisons. We run each experiment with 3 random seeds and report the average performance together with the standard error.}
\vspace{-2mm}

\begin{footnotesize}

\resizebox{0.65\textwidth}{!}{

\begin{tabular}{cccc}
\toprule
& Bitcoin-OTC & Bitcoin-Alpha & UCI-Message\\
\midrule
GCN \cite{kipf2016semi} & 0.0025 & 0.0031 & 0.1141 \\
DynGEM \cite{kamra2017dyngem} & 0.0921 & 0.1287 & 0.1055 \\ 
dyngraph2vecAE \cite{goyal2020dyngraph2vec} & 0.0916 & 0.1478 & 0.0540 \\ 
dyngraph2vecAERNN \cite{goyal2020dyngraph2vec} & \textbf{0.1268} & \textbf{0.1945} & 0.0713 \\ 
EvolveGCN-H \cite{pareja2020evolvegcn} & 0.0690 & 0.1104 & 0.0899 \\ 
EvolveGCN-O \cite{pareja2020evolvegcn} & 0.0968 & 0.1185 & \textbf{0.1379} \\ 
\midrule
ROLAND Moving Average& 0.0468 $\pm$ 0.0022 & 0.1399 $\pm$ 0.0107 & 0.0649 $\pm$ 0.0049 \\ 
ROLAND MLP& 0.0778 $\pm$ 0.0024 & 0.1561 $\pm$ 0.0114 & 0.0875 $\pm$ 0.0110 \\ 
ROLAND GRU& \textbf{0.2203 $\pm$ 0.0167} & \textbf{0.2885 $\pm$ 0.0123} & \textbf{0.2289 $\pm$ 0.0618} \\ 
\midrule
Improvement over best baseline & 73.74\% & 43.33\% & 65.99\% \\ 
\bottomrule
\end{tabular}

}

\label{tab:fixed_split_compare}
\end{footnotesize}
\end{table*}

\section{Experiments}

\subsection{Experimental Setup}

\xhdr{Datasets}
We perform experiments using eight different datasets.
\textbf{(1)} BSI-ZK and \textbf{(2)} BSI-SVT consist of financial transactions among companies using two different systems of the Bank of Slovenia, and each node has five categorical features; each edge has transaction amount as the feature.
\textbf{(3)} Bitcoin-OTC and Bitcoin-Alpha contain who-trusts-whom networks of people who trade on the OTC and Alpha platforms \cite{kumar2018rev2, kumar2016edge}.
\textbf{(4)}  The Autonomous systems AS-733 dataset of traffic flows among routers comprising the Internet \cite{leskovec2005graphs}.
\textbf{(5)} UCI-Message dataset consists of private messages sent on an online social network system among students \cite{panzarasa2009patterns}.
\textbf{(6)} Reddit-Title and \textbf{(7)} Reddit-Body are networks of hyperlinks in titles and bodies of Reddit posts, respectively. Each hyperlink represents a directed edge between two subreddits \cite{kumar2018community}.
Table \ref{tab:datasets} provides summary statistics of the above-mentioned datasets.

\xhdr{Task}
We evaluate the ROLAND framework over the future link prediction task.
At each time $t$, the model utilizes information accumulated up to time $t$ to predict edges in snapshot $t+1$.
We use mean reciprocal rank (MRR) to evaluate candidate models. For each node $u$ with positive edge $(u, v)$ at $t+1$, we randomly sample 1000\footnote{We only sampled 100 negative edges in the BSI-ZK dataset due to memory constraints.} negative edges emitting from $u$ and identify the rank of edge $(u, v)$'s prediction score among all other negative edges.
MRR score is the mean of reciprocal ranks over all nodes $u$.
We consider two different train-test split methods in this paper.
(1) \emph{Fixed-split} evaluates models using all edges from the last 10\% of snapshots.
However, the transaction pattern is constantly evolving in the real world; evaluating models solely based on edges from the last few weeks provides misleading results.
(2) \emph{Live-update} evaluates model performance over all the available snapshots. We randomly choose 10\% of edges in each snapshot to determine the early-stopping condition.
\cut{by randomly sample 10\% of edges from each snapshot as the test edge. Our method leads to a similar amount of test edges but they are distributed across the whole span of the dataset available instead of just the last few snapshots.}

\xhdr{Baselines}
We compare our \name models to 6 state-of-the-art dynamic GNNs.
\textbf{(1)} EvolveGCN-H and \textbf{(2)} EvolveGCN-O
utilizes an RNN to dynamically update weights of internal GNNs, which allows the GNN model to change during the test time \cite{pareja2020evolvegcn}.
\textbf{(3)} T-GCN
internalizes a GNN into the GRU cell by replacing linear transformations in GRU with graph convolution operators \cite{zhao2019t}.
\textbf{(4)} GCRN-GRU and \textbf{(5)} GCRN-LSTM
generalize TGCN by capturing temporal information using either GRU or LSTM. Instead of GCNs, GCRN uses ChebNet \cite{defferrard2016convolutional} for spatial information.
Besides, GCRN uses separate GNNs to compute different gates of RNNs. Hence GCRN models have much more parameters compared with other models \cite{seo2018structured}.
\textbf{(6)} GCRN-Baseline GCRN baseline firstly constructs node features using a Chebyshev spectral graph convolution layer to capture spatial information; afterward, node features are fed into an LSTM cell for temporal information
\cite{seo2018structured}.

\xhdr{ROLAND architecture} \name is designed to re-purpose any static GNN into a dynamic one. Thus the hyper-parameter space of the ROLAND model is similar to the hyper-parameter space of the underlying static GNN.
We use 128 hidden dimensions for node states, GNN layers with skip-connection, sum aggregation, and batch-normalization. We allow for at most 100 epochs for each live-update in each time step before early stopping. In addition, for each type of update-module and dataset, we search the hyper-parameters over (1) numbers of pre-processing, message-passing, and post-processing layers (1 layer to 5 layers); (2) learning rate for live-update (0.001 to 0.01); (3) whether to use single or bidirectional messages; and (4) the $\alpha$ level for meta-learning to determine the best configuration. We report the test set MRR when the best validation MRR is achieved. We use 
NVIDIA RTX 8000 GPU in the experiments.
We implement ROLAND\footnote{The source code of ROLAND is available at \url{https://github.com/snap-stanford/roland}} with the GraphGym library \cite{you2020design}.

\subsection{Results in the Standard Evaluation Settings}
We firstly compare our models with baselines under the standard fixed split setting.
In these experiments, we follow the experimental setting in the EvolveGCN paper \cite{pareja2020evolvegcn}; specifically, we use the same snapshot frequency, set of test snapshots, and the method to compute MRR to ensure fair comparisons.
Table \ref{tab:fixed_split_compare} shows that ROLAND with GRU update-module consistently outperforms baseline models and achieves on average 62.69\% performance gain.
This agrees with the fact that the GRU update-module is more expressive than simple moving average or MLP update-modules. 
The significant performance gain demonstrates that ROLAND can effectively transfer successful static GNN designs to dynamic graph tasks. In Section \ref{subsec:ablation}, we provide comprehensive ablation studies to explain the success of ROLAND further.

\subsection{Results in the Live-update Settings}

\xhdr{Baselines with BPTT training vs. ROLAND}
By default, the baseline models are trained with back-propagation-through-time (BPTT), which requires storing all the historical node embeddings in GPU memory. This training strategy cannot scale to large graphs.
The top part of Table \ref{tab:combined} summarizes the experiments on training the baseline models using BPTT.
Our experiment results indicate that the default BPTT training fails to scale to large datasets, such as BSI-ZK and AS-733, and fails to work with large models such as GCRNs.
For datasets with a moderate number of edges and snapshots like BSI-SVT, BPTT still leads to OOM for large models like GCRNs and TGCN.
Even though this training pipeline works on small datasets, the best model trained from this procedure still underperforms compared to baseline models trained using \name incremental training by almost 10\%.

\xhdr{Baselines with ROLAND training vs. ROLAND}
To quantitatively compare our models with baselines, we re-implemented baseline models and adapted them to be trained using our \name training.
The middle and bottom parts of Table \ref{tab:combined} contrast the performance of our models and baselines on all eight datasets.
All baseline models are successfully trained using our \name training, demonstrating the flexibility of \name
Moreover, ROLAND with GRU update-module outperforms the best baseline model by 15.48\% on average.
Except for the AS-733 dataset, our model consistently outperforms baseline models trained using the ROLAND pipeline. The performance gain of using our models ranges from 2.40\% on BSI-ZK to 44.22\% on the Reddit-Body dataset.
The performance gain is the largest on datasets with rich edge features such as Reddit hyperlink graphs, on which \name brings 38.21\% and 44.22\% performance gains, respectively. These results demonstrate that \name can utilize edge features effectively. We include more analysis of the results in the Appendix.

\begin{table*}[t]
\setlength\tabcolsep{3pt}
\centering

\caption{\textbf{Results in the live-update settings}. We run experiments (except for BSI-ZK) with 3 random seeds to report the average and standard deviation of MRRs. 
We attempted each experiment five times before concluding the out-of-memory (OOM) error.
The top part of the results consists of baseline models trained using BPTT, the middle and bottom portions summarize the performance of baselines, and our models using \name incremental training.}
\begin{footnotesize}
\vspace{-1mm}
\resizebox{0.95\textwidth}{!}{
    \begin{tabular}{ccccccccc}
    \toprule
    &\multicolumn{1}{c}{BSI-ZK}&\multicolumn{1}{c}{AS-733}&\multicolumn{1}{c}{Reddit-Title}&\multicolumn{1}{c}{Reddit-Body}&\multicolumn{1}{c}{BSI-SVT}&\multicolumn{1}{c}{UCI-Message}&\multicolumn{1}{c}{Bitcoin-OTC}&\multicolumn{1}{c}{Bitcoin-Alpha}\\ 
    \midrule 
    \multicolumn{9}{c}{Baseline Models with standard training} \\
    \midrule
    EvolveGCN-H&N/A, OOM&N/A, OOM&N/A, OOM&\textbf{0.148 $\pm$ 0.013}&\textbf{0.031 $\pm$ 0.016}&0.061 $\pm$ 0.040&0.067 $\pm$ 0.035&0.079 $\pm$ 0.032\\
    EvolveGCN-O&N/A, OOM&N/A, OOM&N/A, OOM&N/A, OOM&0.015 $\pm$ 0.006&0.071 $\pm$ 0.009&0.085 $\pm$ 0.022&0.071 $\pm$ 0.025\\
    GCRN-GRU&N/A, OOM&N/A, OOM&N/A, OOM&N/A, OOM&N/A, OOM&0.080 $\pm$ 0.012&N/A, OOM&N/A, OOM\\
    GCRN-LSTM&N/A, OOM&N/A, OOM&N/A, OOM&N/A, OOM&N/A, OOM&\textbf{0.083 $\pm$ 0.001}&N/A, OOM&N/A, OOM\\
    GCRN-Baseline&N/A, OOM&N/A, OOM&N/A, OOM&N/A, OOM&N/A, OOM&0.069 $\pm$ 0.004&\textbf{0.152 $\pm$ 0.011}&\textbf{0.141 $\pm$ 0.005}\\
    TGCN&N/A, OOM&N/A, OOM&N/A, OOM&N/A, OOM&N/A, OOM&0.054 $\pm$ 0.024&0.128 $\pm$ 0.049&0.088 $\pm$ 0.038\\
    \midrule
    \multicolumn{9}{c}{Baseline Models with ROLAND Training} \\
    \midrule
    EvolveGCN-H&N/A, OOM&0.251 $\pm$ 0.079&0.165 $\pm$ 0.026&0.102 $\pm$ 0.010&0.032 $\pm$ 0.008&0.057 $\pm$ 0.012&0.076 $\pm$ 0.022&0.054 $\pm$ 0.015\\
    EvolveGCN-O&0.396&0.163 $\pm$ 0.002&0.047 $\pm$ 0.004&0.033 $\pm$ 0.001&0.018 $\pm$ 0.003&0.066 $\pm$ 0.012&0.032 $\pm$ 0.004&0.034 $\pm$ 0.002\\
    GCRN-GRU&N/A, OOM&\textbf{0.344 $\pm$ 0.001}&0.338 $\pm$ 0.006&0.217 $\pm$ 0.004&0.050 $\pm$ 0.004&0.089 $\pm$ 0.004&0.173 $\pm$ 0.003&0.140 $\pm$ 0.004\\
    GCRN-LSTM&N/A, OOM&0.341 $\pm$ 0.001&0.344 $\pm$ 0.005&0.216 $\pm$ 0.000&0.051 $\pm$ 0.002&0.091 $\pm$ 0.010&0.174 $\pm$ 0.004&\textbf{0.146 $\pm$ 0.005}\\
    GCRN-Baseline&0.754&0.336 $\pm$ 0.002&0.351 $\pm$ 0.001&0.218 $\pm$ 0.002&0.054 $\pm$ 0.002&\textbf{0.095 $\pm$ 0.008}&\textbf{0.183 $\pm$ 0.002}&0.145 $\pm$ 0.003\\
    TGCN&\textbf{0.831}&0.343 $\pm$ 0.002&\textbf{0.391 $\pm$ 0.004}&\textbf{0.251 $\pm$ 0.001}&\textbf{0.157 $\pm$ 0.004}&0.080 $\pm$ 0.015&0.083 $\pm$ 0.011&0.069 $\pm$ 0.013\\
    \midrule
    \multicolumn{9}{c}{ROLAND results} \\
    \midrule
    Moving Average& 0.819& 0.309 $\pm$ 0.011& 0.362 $\pm$ 0.007& 0.289 $\pm$ 0.038& 0.177 $\pm$ 0.006& 0.075 $\pm$ 0.006& 0.120 $\pm$ 0.002& 0.0962 $\pm$ 0.010\\ 
    MLP-Update& 0.834& 0.329 $\pm$ 0.021& 0.395 $\pm$ 0.006& 0.291 $\pm$ 0.008& \textbf{0.217 $\pm$ 0.003} & 0.103 $\pm$ 0.010& 0.154 $\pm$ 0.010& 0.148 $\pm$ 0.012\\ 
    GRU-Update& \textbf{0.851}& \textbf{0.340} $\pm$ 0.001& \textbf{0.425} $\pm$ 0.015 & \textbf{0.362} $\pm$ 0.002 & 0.205 $\pm$ 0.014& \textbf{0.112} $\pm$ 0.008 & \textbf{0.194} $\pm$ 0.004 & \textbf{0.157} $\pm$ 0.007\\ 
    \midrule
    Improvement over & \multirow{2}{*}{2.40\%}&\multirow{2}{*}{-1.16\%}&\multirow{2}{*}{8.70\%}&\multirow{2}{*}{44.22\%}&\multirow{2}{*}{38.21\%}&\multirow{2}{*}{17.89\%}&\multirow{2}{*}{6.01\%}&\multirow{2}{*}{7.53\%} \\
    the best baseline & \\
    \bottomrule
    \end{tabular}
}

\label{tab:combined}
\end{footnotesize}
\end{table*}

\subsection{Ablation Study}
\label{subsec:ablation}

\xhdr{Effectiveness of ROLAND architecture design}
As described in Section 2.2, we introduce several successful GNN designs in static GNNs to ROLAND, so that they can work with dynamic graphs.
Here, we conduct experiments to show that the ROLAND architecture design is effective in boosting performance.

Concretely, we examine the ROLAND framework on the BSI-SVT dataset under the proposed live-update setting. The setting is similar to Table 3 in the main paper.
We found that Batch Normalization, skip-connection, and max aggregation are desirable for GNN architectural design, leading to a significant performance boost. For example, the introduction of skip-connection leads to more than 20\% performance gain.

\begin{figure}[h]
\centering
\includegraphics[width=\linewidth]{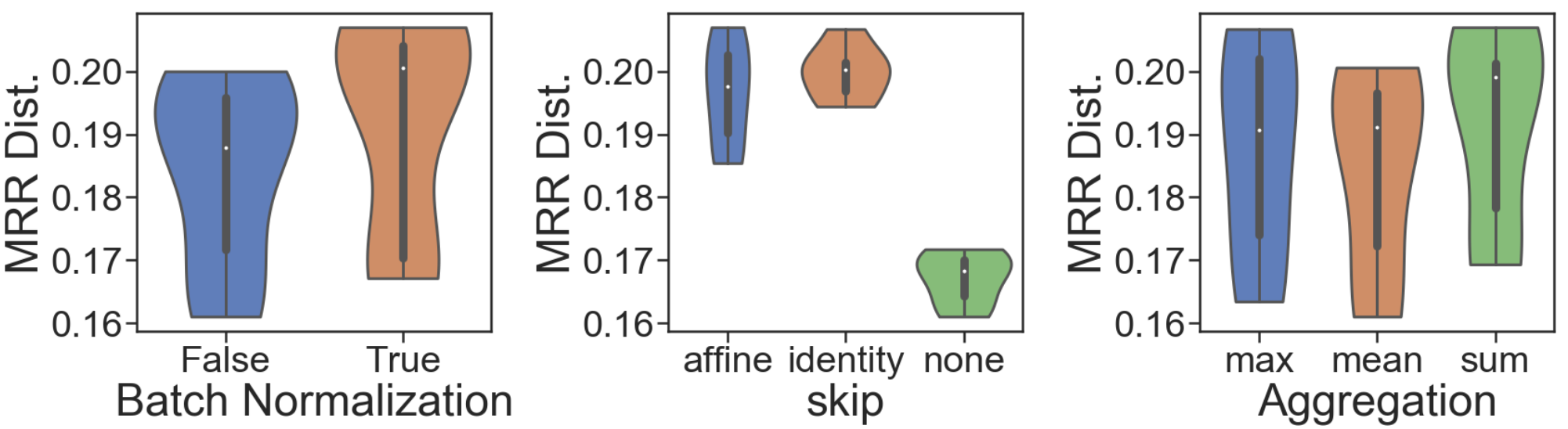} 
\vspace{-4mm}
\caption{\textbf{Effectiveness of ROLAND architectural design}. We run experiments with different GNN models and analyze the effect of differ design dimensions. We show the MRR distribution of all the models under different design options. Results show that batch normalization, skip-connection and max aggregation are desirable for GNN architectural design.}
\label{fig:roland-design-compare}
\end{figure}

\xhdr{Effectiveness of meta-learning}
We examine the effectiveness of meta-learning.
Specifically, for each of three embedding update methods and each dataset, we run 10 experiments with $\alpha \in \{$0.1, 0.2, $\dots$, 1.0$\}$ to study the effect of $\alpha$ on performance\footnote{For BSI-ZK dataset, we run 6 experiments for each update method with $\alpha \in \{0.1,0.3,0.5,0.7,0.9,1.0\}$.}. Here, $\alpha=1$ corresponds to the baseline that directly uses the previous model to initialize the training of $\textsc{Gnn}$.
Table \ref{tab:meta} reports the performance gains in MRR from meta-learning.
Enabling meta-learning leads to various levels of performance gain depending on the dataset and model; for different embedding update designs, the average performance gain ranges from 2.84\% to 13.19\%.
Besides the performance gain in MRRs, we find incorporating meta-learning, in general, improves the stability of performance, as suggested by a lower standard deviation of MRRs. We include the complete results in Appendix.

\begin{figure*}[h]
\centering
\includegraphics[width=0.9\linewidth]{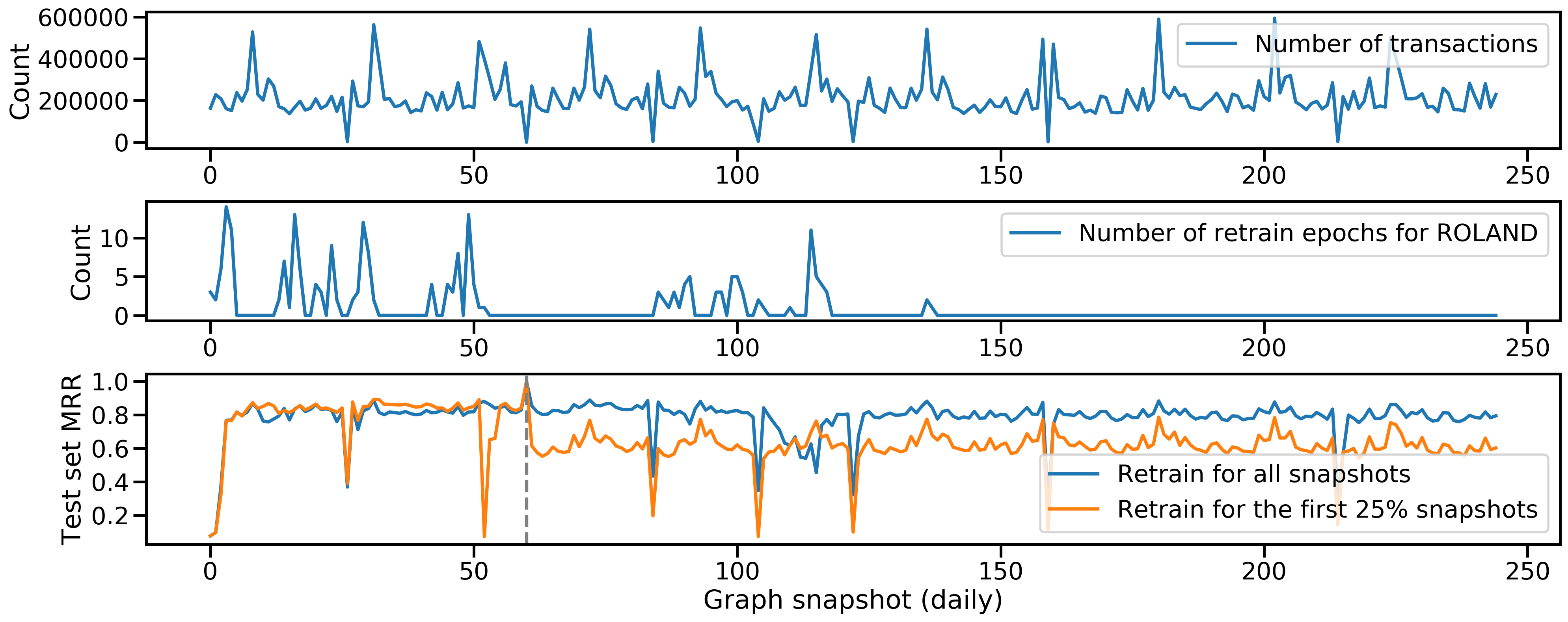} 
\vspace{-1mm}
\caption{\textbf{Effectiveness of ROLAND model retraining} on BSI-ZK dataset. Top: number of transactions in each daily graph snapshot. Middle: number of training epochs before early stopping. Bottom: ROLAND's performance on the test set. We compare the option of retraining for all snapshots (blue curve), with a baseline that stopping retraining after training the first 25\% snapshots (organ curve). }
\label{fig:non-retrain}
\end{figure*}

\xhdr{Effectiveness of model retraining}
We investigate the effectiveness of model retraining in Figure \ref{fig:non-retrain}.
As is shown in Figure \ref{fig:non-retrain}(top), the transaction pattern, measured by the number of transactions, significantly varies over time; in the meantime, ROLAND can automatically retrain itself to fit the data (Algorithm \ref{alg:live-update-evaluation}).
Thanks to the meta-training, ROLAND only requires periodic retraining for a few epochs, and in most snapshots, as is shown in Figure \ref{fig:non-retrain}(middle).
We further compare our strategy with a baseline that stops retraining after training the first 25\% of snapshots. We found that this baseline performs significantly worse than retraining for all the snapshots.

\begin{table*}[ht]
\setlength\tabcolsep{3pt}
\centering
\caption{
\textbf{Ablation study on the effectiveness of meta-learning.}
Except for the BSI-ZK dataset, the results are averaged over 3 random seeds.
``Gain'' refers to the MRR improvement from the best meta-learning setting over the non-meta-learning setting.
}
\vspace{-1mm}
\begin{footnotesize}

\resizebox{\textwidth}{!}{

\begin{tabular}{cccccccccccccccccc}
\toprule
& \multicolumn{2}{c}{BSI-ZK} & \multicolumn{2}{c}{AS-733} & \multicolumn{2}{c}{Reddit-Title} & \multicolumn{2}{c}{Reddit-Body} & \multicolumn{2}{c}{BSI-SVT} & \multicolumn{2}{c}{UCI-Message} & \multicolumn{2}{c}{Bitcoin-OTC} & \multicolumn{2}{c}{Bitcoin-Alpha} & \multirow{2}{*}{Average} \\
\cmidrule(lr){2-3}\cmidrule(lr){4-5}\cmidrule(lr){6-7}\cmidrule(lr){8-9}\cmidrule(lr){10-11}\cmidrule(lr){12-13}\cmidrule(lr){14-15}\cmidrule(lr){16-17}
Model & $\alpha$ & Gain & $\alpha$ & Gain & $\alpha$ & Gain & $\alpha$ & Gain & $\alpha$ & Gain & $\alpha$ & Gain & $\alpha$ & Gain & $\alpha$ & Gain & \\
\midrule
Moving Average & 0.5 & 0.99\% & 0.4 & 4.94\% & 0.7 & 2.68\% & 0.5 & 8.52\% & 0.5 & 3.27\% & 0.7 & 23.73\% & 0.9 & 7.45\% & 0.6 & 6.94\% & 7.33\% \\ 
MLP-Update & 0.5 & 4.04\% & 0.9 & 18.05\% & 0.7 & 1.09\% & 0.4 & 2.51\% & 0.4 & 11.40\% & 0.2 & 27.73\% & 0.3 & 33.76\% & 0.6 & 6.98\% & 13.19\% \\ 
GRU-Update & 0.7 & 0.56\% & 0.5 & 5.15\% & 0.1 & 2.97\% & 0.5 & 8.18\% & 0.8 & 2.10\% & 0.5 & 0.17\% & 0.9 & 2.82\% & 0.8 & 0.77\% & 2.84\% \\ 
\bottomrule
\end{tabular}
}

\label{tab:meta}
\end{footnotesize}
\end{table*}

\section{Additional Related Work}

\xhdr{Dynamic GNNs}
While many works combine GNNs with recurrent models (\eg, GRU cells), we want to emphasize that our ROLAND framework is quite different.
Most existing works take a recurrent model (RNN, Transformer) as the backbone architecture, then (1) replace either the feature encoder \cite{peng2020spatial} with a GNN, or (2) replace the linear layers in the RNN cells with GNN layers \cite{li2018dcrnn_traffic, seo2018structured, zhao2019t}. The first line of approaches ignores the evolution of lower-level node embeddings, while the second approach only utilizes a given GNN layer rather than building upon an established and successful static GNN architecture.
EvolveGCN \cite{pareja2020evolvegcn} proposes to recurrently updates the GNN weights; in contrast, ROLAND recurrently updates hierarchical node embeddings. While the EvolveGCN approach is memory efficient, the explicit historical information is lost (e.g., past transaction information). Our experiments show that EvolveGCN delivers undesirable performance when the number of graph snapshots is large.
Moreover, few existing dynamic GNN works have explored state-of-the-art static GNN designs, such as the inclusion of edge features, batch normalization, skip-connections, etc.
By utilizing mature static GNN designs, we show that ROLAND can significantly outperform existing dynamic GNNs.

\xhdr{Scalable training}
Most existing approaches do not consider scalable training since the current benchmark dynamic graphs are small; thus, the entire graph can often fit into the GPU memory.
Additionally, some dynamic GNN architectures can hardly scale; for example, when using Transformer as the base sequence model requires keeping the entire historical graph in the GPU to compute the attention over time \cite{sankar2020dysat}.
Some existing approaches have used heuristics to scale training. For example, they would limit the number of historical graph snapshots used to predict the current snapshot \cite{zhao2019t}.
In contrast, our proposed incremental training keeps the entire historical information in the node hidden states while only choosing to back-propagate within the given snapshot.

\xhdr{Non-snapshot dynamic graph representation}
Instead of using snapshot-based representation, one can also represent a dynamic graph as a single graph where nodes and edges have time stamps \cite{kumar2019predicting,ma2020streaming,rossi2020temporal}. 
As is described in Section \ref{subsec:preliminaries}, we can easily convert that representation to graph snapshots by grouping all the nodes/edges within a given period into a graph snapshot, and then apply our ROLAND framework.

\xhdr{Other learning methods for dynamic graphs}
Besides GNNs, researchers have explored other learning methods for dynamic graphs, such as matrix factorization based methods \cite{li2017attributed}, random walk based models \cite{nguyen2018continuous,yu2018netwalk}, point process based approaches \cite{trivedi2017know,trivedi2018representation,zhou2018dynamic}. We focus on designing GNN-based methods in this paper due to their performance and inductive learning capabilities.

\section*{Acknowledgements}
We thank Rok Sosic, Daniel Borrajo, Vamsi Potluru, Naren Chittar, Pranjal Patil, Hao Yang, Yuhang Wu, Das Mahashweta for their helpful discussions on the paper. 
We also gratefully acknowledge the support of
Stanford Data Science Initiative and JPMorgan Chase.

\section{Conclusion}
We propose \name, an effective graph representation learning system for building, training, and evaluating dynamic GNNs.
\name helps researchers re-purpose any static GNN for a dynamic graph while keeping effective designs from static GNNs.
\name comes with a live-update pipeline that mimics real-world usage by allowing the model to be dynamically updated while being evaluated.
Our experiments show that dynamic GNNs built using \name framework successfully scale to large datasets and outperform existing state-of-the-art models.
We hope \name can enable more large-scale real-world applications over dynamic graphs.

\bibliography{bibli}
\bibliographystyle{abbrv}

\end{document}